\documentclass[conference]{IEEEtran}
\IEEEoverridecommandlockouts

\usepackage{cite}
\usepackage{amsmath,amssymb,amsfonts}
\usepackage{algorithmic}
\usepackage{graphicx}
\usepackage{textcomp}
\usepackage{xcolor}
\usepackage{booktabs}
\usepackage{multirow}
\usepackage{float}
\def\BibTeX{{\rm B\kern-.05em{\sc i\kern-.025em b}\kern-.08em
		T\kern-.1667em\lower.7ex\hbox{E}\kern-.125emX}}
\begin{document}
	
	\title{Enhancing ECG Classification Robustness with Lightweight Unsupervised Anomaly Detection Filters}
	
	\author{Mustafa Fuad Rifet Ibrahim\textsuperscript{*, 1, 2, 3}\thanks{\textsuperscript{*}~Corresponding author}, Maurice Meijer\textsuperscript{1}, Alexander Schlaefer\textsuperscript{2}, Peer Stelldinger\textsuperscript{3}\\
		\textsuperscript{1}NXP Semiconductors, Eindhoven, The Netherlands\\
		\textsuperscript{2}Institute of Medical Technology and Intelligent Systems, Hamburg University of Technology, Hamburg, Germany\\
		\textsuperscript{3}Faculty of Computer Science and Digital Society, Hamburg University of Applied Sciences, Hamburg, Germany\\
		{\tt\small\{mustafa.ibrahim, maurice.meijer\}@nxp.com}\\
		{\tt\small schlaefer@tuhh.de}\\
		{\tt\small \{mustafa.ibrahim, Peer.Stelldinger\}@haw-hamburg.de}
	}
	
	\maketitle

	\begin{abstract}
Continuous electrocardiogram (ECG) monitoring via wearable devices is vital for early cardiovascular disease detection. However, deploying deep learning models on resource-constrained microcontrollers faces reliability challenges, particularly from Out-of-Distribution (OOD) pathologies and noise. Standard classifiers often yield high-confidence errors on such data. Existing OOD detection methods either neglect computational constraints or address noise and unseen classes separately. This paper investigates Unsupervised Anomaly Detection (UAD) as a lightweight, upstream filtering mechanism. We perform a Neural Architecture Search (NAS) on six UAD approaches, including Deep Support Vector Data Description (Deep SVDD), input reconstruction with (Variational-)Autoencoders (AE/VAE), Masked Anomaly Detection (MAD), Normalizing Flows (NFs) and Denoising Diffusion Probabilistic Models (DDPM) under strict hardware constraints (\(\leq\)512k parameters), suitable for microcontrollers. Evaluating on the PTB-XL and BUT QDB datasets, we demonstrate that a NAS-optimized Deep SVDD offers the superior Pareto efficiency between detection performance and model size. In a simulated deployment, this lightweight filter improves the accuracy of a diagnostic classifier by up to 21.0 percentage points, demonstrating that optimized UAD filters can safeguard ECG analysis on wearables.
	\end{abstract}
	

	\section{Introduction}
	\label{Introduction}
	The transition from controlled datasets to real-world clinical deployment, especially on resource-constrained wearable devices, poses significant challenges. Beyond the fundamental requirements of high accuracy and computational efficiency, a critical challenge arises from the inevitability of encountering Out-of-Distribution (OOD) data. In the context of ambulatory ECG analysis, OOD data typically manifests in two critical forms: ECG signals exhibiting pathologies unseen during the model's training (unseen CVD classes), and signals that are so severely corrupted by noise artifacts (e.g., electrode motion or muscle activity) that they are rendered unsuitable for analysis. Standard classifiers are prone to making erroneous, often high-confidence, predictions on such inputs, which can undermine the reliability of the system \cite{amodei2016concrete}.
	\begin{figure}[t]
		\centering
		\includegraphics[width=0.48\textwidth]{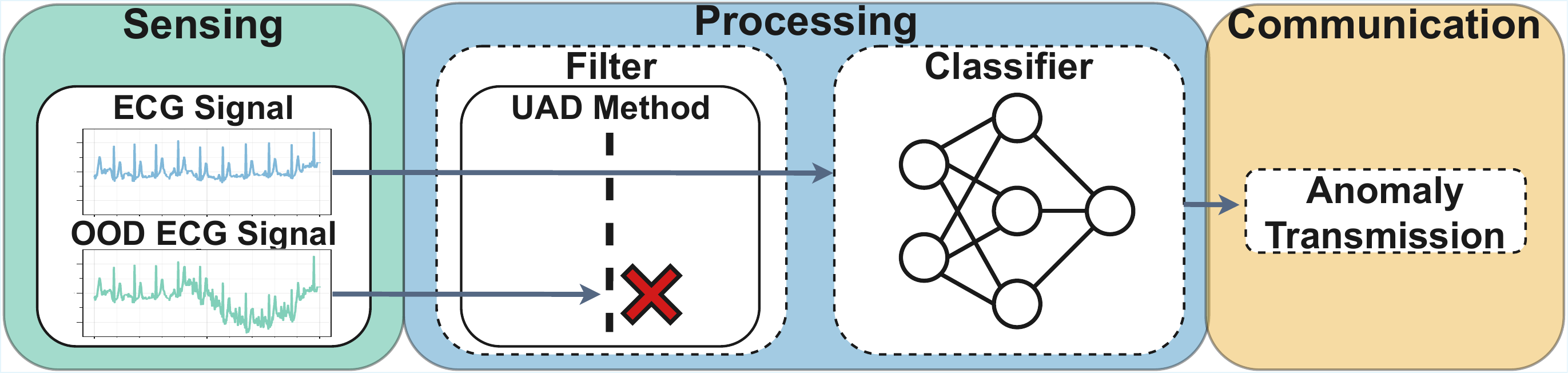}
		\caption{Schematic overview of the proposed robust ECG monitoring architecture. An independent UAD filter is deployed upstream of the diagnostic classifier. The filter assesses incoming ECG signals. If they are deemed anomalous (OOD or unsuitable for analysis due to noise), the signal is rejected, thereby protecting the classifier from unreliable inputs and improving overall system robustness.}
		\label{fig:system_concept}
	\end{figure}
	Addressing this challenge through purely supervised learning is impractical, as it requires an exhaustive, labeled dataset encompassing all possible CVDs and noise profiles encountered in daily life. Consequently, research has focused on methods for OOD detection. Existing approaches like uncertainty quantification (UQ) (e.g., using MC Dropout or Deep Ensembles) \cite{barandas2024evaluation} can detect OOD data but suffer from inefficiency due to multiple ensemble members of forward passes. Furthermore, they frequently address the detection of unseen classes and the identification of noise in isolation.
	
	To address these gaps, this paper investigates the application of Unsupervised Anomaly Detection (UAD) as an independent filtering mechanism deployed upstream of a diagnostic classifier, as illustrated in \figurename\ \ref{fig:system_concept}. Our approach represents a single-inference architecture constrained to under 512k parameters, ensuring compatibility with the limited flash memory and SRAM of low-power microcontrollers. Since the filter can run independently from the classifier, the classifier will not be used when inputs are filtered out, further saving energy. UAD methods learn a compact representation of the in-distribution data without requiring labeled examples of anomalies, enabling the detection of significant deviations caused by either novel pathologies or noise. We conduct a systematic evaluation of diverse UAD methodologies, including Deep Support Vector Data Description (Deep SVDD), reconstruction-based methods (autoencoders (AE) and variational autoencoders (VAE)), Masked Anomaly Detection (MAD), Denoising Diffusion Probabilistic Models (DDPM), and normalizing flows (NF). Our analysis focuses specifically on the trade-off between detection performance and computational efficiency essential for deployment in resource-constrained environments. Our contributions are as follows: (i) We benchmark six distinct UAD methodologies, optimized via Neural Architecture Search (NAS) under strict resource constraints (\(\le\)512k parameters) and evaluate their efficacy for the detection of OOD CVD classes and signals unsuitable for analysis due to noise using the PTB-XL and BUT QDB datasets; (ii) We empirically identify Deep SVDD as the superior methodology for this application, demonstrating that it offers the optimal balance between computational efficiency and detection performance; (iii) We validate the efficacy of an integrated classifier-filter system in a realistic deployment simulation incorporating unseen CVD classes and calibrated real-world ambulatory noise. We show that the upstream UAD filter significantly enhances diagnostic robustness, yielding an accuracy improvement of up to 21.0 percentage points over a classifier-only baseline.
	
	\section{Related Work}
	\label{RelatedWork}
	The deployment of ML models for ECG analysis in real-world clinical settings introduces significant challenges beyond controlled environments. In this section we review existing methods addressing the challenges most relevant to our work.
	
	The reliability of automated ECG analysis is heavily dependent on input signal quality, as noise artifacts (e.g., muscle activity, electrode motion) prevalent in ambulatory monitoring can severely impair diagnostic accuracy. The literature addresses this through Signal Quality Assessment (SQA) and Uncertainty Quantification (UQ). SQA methods explicitly classify noise types \cite{satija2017automated} while UQ indirectly flags low-confidence noisy inputs, for example through predictive entropy \cite{jahmunah2023uncertainty}. Unlike these, our UAD explicitly rejects unsuitable signals regardless of noise type.
	
	UAD methods identify deviations from a learned distribution of normal data without labeled anomalies. In ECG analysis, UAD has been explored primarily to differentiate abnormal patterns from normal sinus rhythm or to enhance supervised learning. Atamny et al. \cite{atamny2023outlier} conducted a comparative study of various unsupervised models, including AEs, VAEs, diffusion models, NFs, and Gaussian Mixture Models (GMMs) to differentiate abnormal ECG signals from healthy ones. In a different approach, Jiang et al. \cite{jiang2024self} employed self-supervised anomaly detection as a pretraining mechanism to improve generalization to rare classes. In contrast, our work concentrates on deploying UAD as an independent, upstream filter designed to reject entirely unseen disease classes and signals rendered unsuitable for further analysis by noise.
	
	The literature most closely related to our research involves detecting OOD samples corresponding to unseen CVDs, primarily through UQ techniques or post-hoc analysis of supervised classifiers. Barandas et al. \cite{barandas2024evaluation} evaluated UQ methods (Deep Ensembles, MC Dropout) in a multi-label ECG setting. They assessed OOD detection using the PTB-XL dataset by treating specific superclasses (myocardial infarction and hypertrophy) as OOD, similar to our design. However, their focus was on UQ applied to supervised classifiers rather than independent UAD filters, and they did not explicitly address the simultaneous detection of signals unsuitable for analysis due to noise. Elul et al. \cite{elul2021meeting} employed a multi-head architecture where an unknown class is inferred if all binary classifiers output a negative prediction, supplemented by MC Dropout for uncertainty estimation. While addressing OOD detection, their strategy is tied to the classifier's architecture, not a dedicated UAD filter, and overlooks efficiency constraints crucial for wearables. Yu et al. \cite{yu2025trustworthy} proposed a trustworthy diagnosis method using post-hoc Energy and ReAct techniques on a CNN-Attention classifier to recognize OOD heart diseases. Eidheim \cite{eidheim2025quantifying} utilized latent space features from a trained supervised classifier (e.g., xResNet1D101) and applied traditional anomaly detection methods (e.g., Mahalanobis Distance) to these features. Similar to UQ approaches, these methods rely on analyzing the supervised model's behavior or features.
	
	In summary, while prior work has made significant strides in OOD detection and UQ for ECG analysis, these efforts often rely on supervised features, address unseen classes and noise in isolation, or overlook the efficiency constraints required for resource-limited environments. Our work addresses these gaps by systematically evaluating and optimizing a diverse range of UAD methods for the simultaneous detection of both unseen CVDs and ECG samples unsuitable for analysis due to noise, focusing specifically on the performance-efficiency trade-off essential for deployment on wearable devices.
	
	\section{Methodology}
	\label{Methodology}
	This section details the datasets utilized, the UAD methods investigated, and the experimental protocols designed to evaluate their performance in detecting both OOD CVDs and ECG signals unsuitable for analysis due to noise.
	
	\subsection{Datasets and Preprocessing}
	
	We used three datasets in this study. PTB-XL is a large, publicly accessible clinical ECG database comprising 21,799 12-lead ECG recordings, each 10 seconds in duration, from 18,869 patients \cite{PTB-XL:Website}. Diagnostic labels include five superclasses: Normal (NORM), Myocardial Infarction (MI), Conduction Disturbance (CD), ST/T Change (STTC), and Hypertrophy (HYP). There are potentially multiple superclasses per record, as they are not mutually exclusive. We use the recommended stratified 10-fold cross-validation splits and the 500 Hz ECG signals. The Brno University of Technology ECG Quality Database (BUT QDB) is designed for evaluating signal quality algorithms \cite{BUT-QDB:Website}. It contains 18 long-term, single-lead ECG recordings (sampled at 1,000 Hz) collected during normal daily activities. Expert annotations classify signal quality into three categories: Class 1 (clear visibility of QRS complex, P waves and T waves), Class 2 (only QRS complexes reliably detectable), and Class 3 (signal unsuitable for analysis, QRS complexes undetectable). The MIT-BIH Noise Stress Test database provides realistic noise recordings characteristic of ambulatory ECGs \cite{moody1984noise}. It includes three half-hour recordings, among which is the electrode motion artifact ('em') record, utilized in this study for noise injection, that contains significant electrode motion artifacts along with baseline wander and muscle noise. While PTB-XL is a clinical resting dataset, it contains a diverse set of pathologies. By injecting calibrated noise from ambulatory ECGs into the clinical signals, we synthesize the signal degradation characteristic of wearable environments while retaining the ground-truth pathological labels required for rigorous OOD evaluation. All ECG recordings were segmented into 10-second windows. These windows were resampled to a uniform length of 512 timesteps, a configuration determined during pre-screening tests to optimize the balance between computational efficiency and model performance. For the PTB-XL dataset, z-score normalization was applied, as it yielded superior performance. For the BUT QDB dataset, instance normalization proved more effective.
	
	\subsection{Unsupervised Anomaly Detection Methods}
	
	We investigated six UAD methods for detecting anomalous ECG signals. To optimize hyperparameters and evaluate the performance-efficiency trade-off, a NAS using random search (100 trials) was conducted for each method. Models were constrained to a maximum of 512k parameters to ensure suitability for resource-constrained microcontrollers. While a floating-point 32-bit model of this size ($\sim$2MB) would saturate the flash memory of typical low-power MCUs (e.g., STM32U3 \cite{STM32}, NHS52S04 \cite{NHS52S04}), this search space allows us to accomodate models deployed with 8-bit quantization (reducing footprint to $\sim$0.5MB) and to identify the point of diminishing returns where added complexity no longer yields detection gains. Deep SVDD aims to map normal data into a compact hypersphere of minimum volume \cite{ruff2018deep}. The anomaly score is the squared Euclidean distance to the center in the output space. A 1D ResNet architecture was employed. The NAS optimized the number of layers (2-8), filters (4-1024), kernel sizes (3-11), strides (1-2), and the latent dimension size (8-1024).
	
	MAD utilizes a self-supervised approach similar to masked autoencoders \cite{fu2022mad}. During training, 5\% of time steps (determined via pre-screening) are masked with random values, and the model reconstructs the original values using bidirectional context. The anomaly score is the sum of reconstruction errors calculated by sequentially masking and reconstructing each time step during inference. A standard transformer architecture with an initial patching layer was used. The NAS varied the number of transformer blocks (1-10), hidden dimensions (32-512), attention heads (1-32), linear layer sizes (32-2048), patch size (4-64), and patch overlap (0-75\%).
	
	For traditional reconstruction-based methods we used AEs and VAEs to reconstruct given inputs. We utilized the L2 reconstruction error as the anomaly score. Stacked 1D convolution layers were used for both encoder and decoder architectures. The NAS optimized the number of layers (2-8), filters (4-1024), kernel sizes (3-11), strides (1-2), and latent dimension size (12-1024). For the VAE, the KL loss weight was also varied (0.1-1.0).
	
	To use DDPMs for anomaly detection, we adopted the approach of Wyatt et al. \cite{wyatt2022anoddpm}: an anomalous input is partially diffused and then reconstructed. The model, trained only on the normal data manifold, attempts to "repair" abnormal regions during denoising. The reconstruction error serves as the anomaly score. A 1D U-Net architecture with convolution and attention layers was implemented using the denoising-diffusion-pytorch package \cite{denoisingdiffusionpytorch}. The NAS varied the diffusion steps for the noising process (100-1000) and denoising process (10-1000), U-Net architecture parameters (layers (1-10), initial filter (12-256), filter expansion factor (1-8), attention heads (2-8), hidden dimensions (12-256)), and the U-Net objective (predicting noise, the original input or a variable from angular parametrization as in Salimans and Ho \cite{salimans2022progressive}).
	
	NFs are exact-likelihood generative models that learn complex data distributions by transforming a simple base distribution through invertible mappings. We implemented a multiscale GLOW architecture \cite{kingma2018glow} with affine coupling layers and invertible 1x1 convolutions, utilizing the normflows package \cite{Stimper2023}. The exact probability density of a sample serves as the anomaly score. Stacked 1D convolutions were used for the subnetwork within the coupling layers. The NAS optimized the subnetwork architecture (layers (1-5), filters (4-1024), kernel sizes (3-11)) and the flow architecture (number of levels (1-5), coupling blocks (1-10) , split ratios (0.25-0.75) and squeeze factors (1-4)).

	\subsection{Experimental Setup}
	
	Two main experiments were conducted to assess the UAD methods and the overall system performance. During the NAS phase (Experiment 1), each configuration was trained three times to account for initialization variability. For final testing (Experiments 1 and 2), evaluations were repeated five times, reporting the mean and standard deviation of the relevant metrics.
	
	\begin{figure*}[ht]
		\centering
		\includegraphics[width=0.9\textwidth]{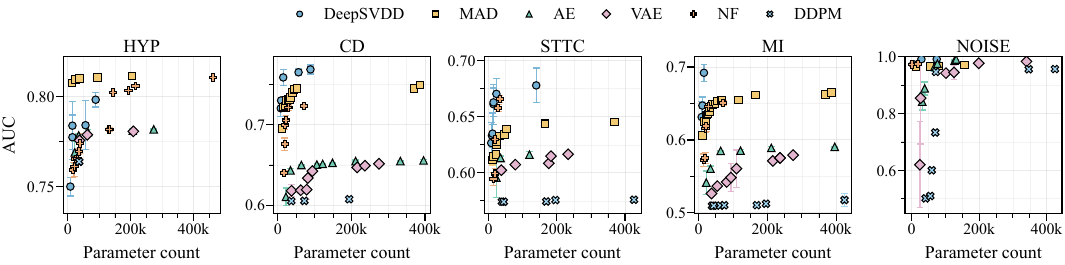}
		\caption{Performance-efficiency trade-offs (Pareto fronts) resulting from the NAS in Experiment 1 across five anomaly detection scenarios. Each panel illustrates the trade-off between detection performance, measured by AUC (higher is better), and computational efficiency, measured by parameter count (lower is better). The upper-left region represents the optimal trade-off. Deep SVDD, NF and MAD consistently demonstrate superior efficiency compared to AE, VAE and DDPM in OOD CVD detection tasks. All methods achieve very high AUC for the detection of ECG signals unsuitable for analysis due to noise (rightmost panel).}
		\label{fig:uad_res}
	\end{figure*}
	
	\subsection*{Experiment 1 - Detection Performance of OOD CVD Classes and Signals Unsuitable for Analysis due to Noise}
	The first experiment evaluated the performance of the UAD methods across two distinct tasks: Detection of ODD classes and of signals unsuitable for analysis due to noise. Using the PTB-XL dataset, four scenarios were created. In each scenario, one diagnostic superclass (MI, CD, STTC, or HYP) was designated as the OOD class. All samples containing the OOD class, including those where it co-occurred with in-distribution classes, were removed from the training data (folds 1-8) but not from the validation (fold 9) and test (fold 10) sets. Performance was measured using the Area Under the Receiver Operating Characteristic curve (AUC) to get a threshold independent assessment. Using the BUT QDB dataset, the data was split patient-wise to avoid data leakage. The training set included patients with only Class 1 and Class 2 noise labels. The test set included patients exhibiting Class 3 noise. A 10-second window was labeled as unsuitable for analysis due to noise (anomalous) if at least 1 second within that window was annotated as Class 3.

	\subsection*{Experiment 2 - Integrated Classifier-Filter System Performance}
	The second experiment was designed to assess the practical benefit of employing a UAD filter upstream of a standard multilabel classifier in a realistic deployment scenario involving both OOD classes and signals unsuitable for analysis due to noise. A high-performing multilabel classification model was required. We selected the resnet1d\_wang model, identified as the best-performing architecture on the PTB-XL diagnostic superclasses in the comprehensive benchmark by Strodthoff et al. \cite{Strodthoff:2020Deep}. The classifier was trained in the same four scenarios as Experiment 1, using only the four in-distribution classes and excluding the OOD class from training to assess in-distribution performance. Deep SVDD was selected as the filter mechanism, as it demonstrated the best average performance across both detection of OOD classes and of signals unsuitable for analysis due to noise in Experiment 1. The hyperparameter configuration yielding the top result for each specific scenario was utilized.
	To simulate realistic conditions, the PTB-XL validation and test sets were modified to include calibrated amounts of noise using the nst script from PhysioNet \cite{NST-script}. The 'em' noise record from the MIT-BIH Noise Stress Test Database was injected. The Signal-to-Noise Ratio (SNR) used by nst is defined based on the peak-to-peak amplitude of QRS complexes. We determined the appropriate SNR for injection by estimating the noise level present in real-world signals unsuitable for analysis (BUT QDB Class 3 samples). For this, we used wavelet denoising with Donoho's universal threshold and soft thresholding \cite{donoho2002noising}.This calibrated noise was applied to approximately 16.5\% of the validation and test data, matching the prevalence of Class 3 noise windows (signals unsuitable for analysis) observed in our BUT QDB test set. Crucially, this injection ratio was applied across both the in-distribution and OOD samples within each scenario (HYP, CD, STTC, MI) to ensure a sufficient representation of samples that are both unsuitable for analysis due to noise and belong to an unseen class.
	A custom accuracy metric was defined to evaluate the combined system, accounting for both filtering and classification efficacy. An outcome was correct if valid and classified correctly, or if invalid (OOD CVD or Noisy) and rejected by the filter. The performance of the classifier-only approach and the integrated classifier-filter system were compared using this custom accuracy. The impact of the filter was further analyzed by plotting the custom accuracy against the rejection rate.
	
	\section{Results}
	\label{Results}
	This section presents the empirical results of the experiments designed to evaluate the efficacy of various UAD methods for identifying OOD CVDs and ECG signals unsuitable for analysis due to noise, and the subsequent impact of integrating the best-performing method into a classification pipeline.
	
	\subsection{Experiment 1 - Detection Performance of OOD CVD Classes and Signals Unsuitable for Analysis due to Noise}
	\figurename\ \ref{fig:uad_res} illustrates the Pareto fronts resulting from the NAS, depicting the trade-off between AUC and parameter count for each method and scenario. A general trend across the OOD scenarios (HYP, CD, STTC, MI) is that performance initially increases with model complexity before reaching a plateau, often well within the parameter constraint. The results reveal significant differences in performance profiles among the UAD methods. Deep SVDD, NF, and MAD consistently demonstrated superior efficiency, generally occupying the upper-left regions of the Pareto fronts. This indicates higher AUC scores with fewer parameters compared to traditional reconstruction-based approaches (AE, VAE) and DDPM, which were often dominated across the range of parameter counts in OOD tasks.
	
		\begin{table*}[h]
		\caption{Summary of the best detection performance (AUC) achieved by each UAD method during NAS (Experiment 1), constrained to 512k parameters. Results are reported as mean \(\pm\) standard deviation over five runs. For each scenario, the top-performing method and any method within one standard deviation are highlighted in bold font.}
		\begin{center}
			\begin{tabular}{@{}lccccc@{}}
				\toprule
				\multicolumn{1}{c|}{\multirow{2}{*}{\begin{tabular}[c]{@{}c@{}}UAD\\ Method\end{tabular}}} & \multicolumn{5}{c}{AUC}                                                                                                                 \\ \cmidrule(l){2-6} 
				\multicolumn{1}{c|}{}                                                                      & HYP                       & CD                       & STTC                       & MI                       & Noise                    \\ \midrule
				Deep SVDD                                                                                  & 0.798 \(\pm\) 0.004           & \textbf{0.769 \(\pm\) 0.006} & \textbf{0.678 \(\pm\) 0.015}   & \textbf{0.692 \(\pm\) 0.012} & \textbf{0.992 \(\pm\) 0.002} \\ \midrule
				MAD                             & \textbf{0.811 \(\pm\) 0.001}  & 0.749   \(\pm\) 0.002        & 0.645   \(\pm\) 0.001          & 0.665   \(\pm\) 0.003        & 0.971 \(\pm\) 0.002          \\ \midrule
				AE                                                                                & 0.782 \(\pm\) 0.0004          & 0.656   \(\pm\) 0.001        & 0.616   \(\pm\) 0.003          & 0.591\(\pm\)   0.001         & \textbf{0.991 \(\pm\) 0.001} \\ \midrule
				VAE                         & 0.781 \(\pm\) 0.001           & 0.652   \(\pm\) 0.001        & 0.617   \(\pm\) 0.002          & 0.579   \(\pm\) 0.001        & 0.984 \(\pm\) 0.002          \\ \midrule
				NF                                & \textbf{0.811 \(\pm\) 0.0003} & 0.723   \(\pm\) 0.0003       & \textbf{0.666   \(\pm\) 0.003} & 0.650   \(\pm\) 0.004        & 0.975 \(\pm\) 0.001          \\ \midrule
				DDPM                                                                                  & 0.764 \(\pm\) 0.001           & 0.607   \(\pm\) 0.003        & 0.576   \(\pm\) 0.001          & 0.518   \(\pm\) 0.009        & 0.957 \(\pm\) 0.001          \\ \bottomrule
			\end{tabular}
			\label{tabRes}
		\end{center}
	\end{table*}
	
	\begin{figure}[t]
		\centering
		\includegraphics[width=0.35\textwidth]{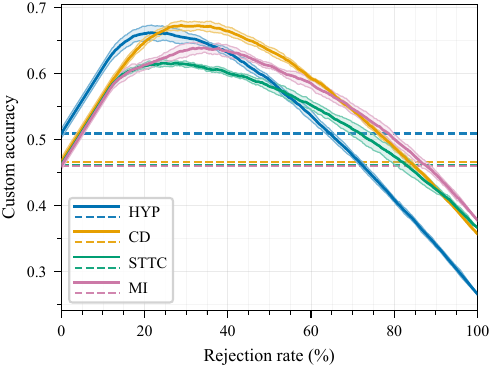}
		\caption{System performance (custom accuracy) as a function of the rejection rate across the four OOD scenarios. The rejection rate (x-axis) is controlled by varying the decision threshold of the Deep SVDD filter. A stricter threshold yields a higher rejection rate. Dashed horizontal lines represent the baseline accuracy of the classifier-only system (0\% rejection rate). The curves demonstrate a critical trade-off. Accuracy initially increases as anomalies are correctly filtered out, peaks at an optimal operating point and subsequently declines as the filter becomes overly aggressive and rejects in-distribution data.}
		\label{fig:acc_res}
	\end{figure}
	
	The difficulty of the detection task also varied significantly depending on the scenario. Detecting HYP as OOD was the most tractable OOD task, with two methods achieving AUC scores above 0.80. Conversely, detecting STTC and MI proved more challenging, with peak AUC scores generally remaining below 0.70. In contrast, the detection of ECG signals unsuitable for analysis due to noise (NOISE panel in \figurename\ \ref{fig:uad_res}) was highly successful across all methods. Nearly all evaluated architectures achieved AUC scores approaching 1.0, irrespective of the parameter count, indicating that severely corrupted ECG signals are easily distinguishable from acceptable quality signals by these UAD approaches.
	
	\tablename\ \ref{tabRes} summarizes the best-performing AUC results achieved during the NAS for each UAD method and scenario. Deep SVDD demonstrated the most robust and consistent high performance across the varied OOD tasks. It achieved the highest AUC in three of the four OOD scenarios: CD (0.769 \(\pm\) 0.006), STTC (0.678 \(\pm\) 0.015), and MI (0.692 \(\pm\) 0.012). In the HYP scenario, MAD (0.811 \(\pm\) 0.001) and NF (0.811 \(\pm\) 0.0003) achieved the top result, marginally outperforming Deep SVDD (0.798 \(\pm\) 0.004). Traditional reconstruction-based methods (AE, VAE) and DDPM generally underperformed in OOD detection; for example, in the MI scenario, the AUC for DDPM was only 0.518 \(\pm\) 0.009.
	
	While all evaluated methods successfully identified signals unsuitable for analysis due to noise (AUC \(\ge\) 0.957), differentiating unseen pathologies revealed distinct architectural biases. AE and VAE models likely underperformed because they generalized too well, accurately reconstructing subtle anomalies rather than flagging them \cite{bouman2025autoencoders}. MAD showed promise for local violations (e.g., Hypertrophy strain patterns \cite{hancock2009aha}) but likely struggled when widespread anomalies corrupted the bidirectional context required for accurate infilling. Normalizing Flows appeared susceptible to assigning high likelihoods to OOD data based on plausible local correlations, failing to capture global structural defects \cite{schirrmeister2020understanding, kirichenko2020normalizing}. Finally, DDPMs exhibited the weakest performance. We hypothesize that standard white Gaussian noise failed to sufficiently corrupt the dominant low-frequency components of the ECG, allowing anomalies to survive the denoising process \cite{wyatt2022anoddpm}.
	
	The results in \tablename\ \ref{tabRes} also reveal significant variability in detecting different diagnostic superclasses as OOD. Detecting MI and STTC proved substantially more challenging than HYP or CD. We hypothesize that this disparity is attributable mostly to the effects of clinical confounding where in-distribution pathologies mimic the OOD class. Abnormalities during the ventricular repolarization phase (ST segment and T wave) are not exclusive to STTC or MI; they frequently occur as secondary effects of HYP and CD. In the context of UAD, this confounding is critical. When STTC or MI is held out, the training data still contains samples with HYP and CD. Consequently, the UAD model learns that the morphological features associated with secondary effects of HYP and CD are part of the in-distribution manifold. When subsequently presented with STTC or MI, the model struggles to identify them as anomalous because similar morphologies are already represented within its learned boundary of normality, directly accounting for the lower AUC scores observed.
	
	Given its superior average performance across both detection of OOD classes and of signals unsuitable for analysis due to noise, Deep SVDD was selected as the filter mechanism for the subsequent experiment. The best parameter selection in terms of AUC from the NAS for each OOD type was as follows: \textbf{CD and HYP}: layers: 3, filter\_0: 12, kernel\_0: 11, stride\_0: 1, filter\_1: 24, kernel\_1: 9, stride\_1: 1, filter\_2: 128, kernel\_2: 3, stride\_2: 1, latent dim.: 128; \textbf{STTC}: layers: 3, filter\_0: 128, kernel\_0: 3, stride\_0: 2, filter\_1: 128, kernel\_1: 3, stride\_1: 1, filter\_2: 32, kernel\_2: 3, stride\_2: 1, latent dim.: 128; \textbf{MI}: layers: 3, filter\_0: 16, kernel\_0: 9, stride\_0: 1, filter\_1: 16, kernel\_1: 5, stride\_1: 1, filter\_2: 16, kernel\_2: 5, stride\_2: 1, latent dim.: 512.

	\subsection{Experiment 2 - Integrated Classifier-Filter System Performance}
	\figurename\ \ref{fig:acc_res} illustrates the relationship between the system's rejection rate and the resulting custom accuracy. The rejection rate (x-axis) is varied by adjusting the anomaly score threshold of the Deep SVDD filter. A stricter threshold increases the filter's sensitivity, thereby increasing the rejection rate. The dashed horizontal lines indicate the baseline accuracy of the classifier-only system, corresponding to a 0\% rejection rate. In all scenarios, as the rejection rate increases from zero, the custom accuracy rises sharply, significantly surpassing the baseline performance by up to 21.0 percentage points (CD scenario). This improvement is attributed to the successful filtering of OOD and noisy samples that the classifier would otherwise misclassify. However, the curves demonstrate a clear trade-off between robustness and utility. The accuracy peaks at an optimal rejection rate, typically between 20\% and 40\% across the different scenarios. In this optimal region, the benefit of correctly rejecting anomalies significantly outweighs the cost of occasionally rejecting in-distribution data. Beyond this optimum, the accuracy declines because the filter becomes overly aggressive, increasingly rejecting in-distribution samples that the classifier could have correctly processed, thus diminishing the overall utility of the system.

	\section{Conclusion}
	\label{Conclusion}
	
	This study demonstrated that deploying a lightweight UAD filter upstream of an ECG classifier significantly enhances robustness. By benchmarking six architectures under strict constraints (\(\leq\)512k parameters), we identified Deep SVDD as the Pareto-optimal solution. In a realistic deployment simulation, this filter rejected unreliable inputs—both noise and unseen pathologies—improving diagnostic accuracy by up to 21.0 percentage points. Future work will focus on implementing the models on physical microcontrollers to measure real-world latency and power consumption, exploring options to improve UAD performance by e.g. using high-level semantic representations for NFs or utilizing simplex noise fo DDPMs, and incorporating multimodal data (e.g. accelerometer data) to differentiate between motion artifacts and physiological anomalies and thus allow different system responses in those two cases.
	
	\section*{Acknowledgment}
	We thank the Bundesministerium für Forschung, Technologie und Raumfahrt (BMFTR) for their support under the project 16KISK221 ("Holistische Entwicklung leistungsfähiger 6G-Vernetzung für verteilte medizintechnische Systeme (6G-Health)").

	\section*{Code Availability}
	The code supporting the findings of this study will be made publicly available upon publication.

	\bibliographystyle{./IEEEtran}
	\bibliography{./IEEEabrv,./IEEEexample}
	
\end{document}